\documentclass{article}
\usepackage{graphicx,subcaption,natbib}
\usepackage[margin=1in]{geometry}
\usepackage{sgamex}
\newtheorem{story}{Story}

\title{Is Recursive Self-Improvement Safe?}
\author{Vincent Conitzer}

\usepackage{ifthen,xcolor}
\newboolean{commentsactivated}
\setboolean{commentsactivated}{false}
\newcommand{\vc}[1]{\ifthenelse{\boolean{commentsactivated}}{{\color{olive} {\em Vince: #1 }}}{}}

\begin{document}

\maketitle

\begin{abstract}
The phrase {\em recursive self-improvement (RSI)} refers to the idea of an AI system designing and creating another, better AI system, which in turn designs and creates an even better one, and so on.  While the idea has a long history, it has always been an abstract, futuristic idea -- until recently.  Today's frontier AI has attained a set of abilities, including in programming, mathematics, research, tool use, and working coherently over extended periods of time, that make RSI seem plausible.  AI companies are starting to pursue it, but many of the involved people are also expressing concerns, asking whether humanity should really set this in motion and under what conditions.  I will 
discuss the situation, argue that we do not even have clear frameworks for thinking about the likely outcomes of RSI, and do a quick game-theoretic analysis of how to choose whether to initiate RSI when another party is making the same choice.
\end{abstract}

\section{Introduction}

The idea of recursive self-improvement is a relatively old one. It dates back at least to I.~J.~Good:

\begin{quote}
[...] an ultraintelligent machine could design even better machines; there would then unquestionably be an ``intelligence explosion,'' and the intelligence of man would be left far behind [...] \citep{Good65:Speculations}.
\end{quote}
For over half a century, it remained a theoretical, futuristic idea. 
Stuart Russell, in his 2020 book {\em Human Compatible}, has a short section on RSI, concluding with (in the context of analyzing what an AI is really pursuing as its purpose): 

\begin{quote}
AI researchers are only just beginning to get a handle on how to analyze even the simplest kinds of real-decision making systems, let alone machines intelligent enough to design their own successors. We have work to do.~\citep{humancompatible}
\end{quote}
But now, in 2026 -- even though I'm not sure we have all that much more of a handle on analyzing those simpler systems --
we have arrived at the point that RSI is considered a real possibility.
One company, appropriately called {\em Recursive}, is all in on  RSI~\citep{Knight26:Recursive}.  Their website proudly states: ``We are researchers and builders pioneering safe, self-improving superintelligence that expands the boundaries of human possibility.''  We'll come back to the ``safe'' part of that statement later in this article.  Other companies in the space include {\em Ricursive Intelligence} (that's not a typo), {\em Mirendil}, and {\em Discovery Loop}.

But it's not just some wide-eyed new startups; RSI is also very much on the mind of the major frontier AI companies. 
Jacob Coxon, having worked at both OpenAI and Anthropic, resigned from the latter because, he posted, both companies ``are racing straight to self-improving superintelligence and gambling with our lives''; his tweet went viral, bringing such questions to a broader audience~\citep{Bellan26:Gambling}.

This quick transition, where within a few years RSI went from an old theoretical idea of how humanity might finally be left in the dust, to the hot new technique in Silicon Valley that is scaring the world, is so jarring as to defy belief.  Surely it's just over-the-top hype?  Or if it is not, surely what they are actually talking about is not the thing that brings about humanity's successor?

Unfortunately I don't think the answers are that simple or reassuring.  In this article, I'll lay out what I think of it all.

\section{Is it all just hype?}

It is generally natural, and good, to be skeptical about companies' claims about how amazingly powerful their technology is and will soon become, especially when some of these companies are likely to have IPOs at extremely high valuations soon.  When people in the frontier AI companies express concern about the {\em risks} of such powerful technology, similar skepticism is  often used to downplay these warnings -- ``They're just hyping their products'' -- though one has to wonder whether any company would really adopt a marketing strategy based on how damaging its product can be.\footnote{My personal impression is that many of the individual employees who are calling out the risks are entirely sincere in their concerns, but I won't base what follows on that assessment.}

In any case, after a number of hacking events, the risks of today's AI to cybersecurity are now well established.  Interpretations of these risks still vary, in particular in how much the AI itself is responsible versus the humans that instruct them and place them in environments.  One may also take a more positive, defensive view of being able to use frontier AI to identify and patch all sorts of security holes.  But that the technology poses significant cybersecurity risk is at this point undeniable.  

In contrast, for RSI, there is less direct evidence.  We see AI succeed at mathematical and programming challenges, write research papers from scratch (even if the writing is often still poor, especially when it comes to making the content understandable to a human reader), and generally exhibit broad capabilities.  I have a hobby of trying to get silly responses out of the models, but succeeding at that with the {\em leading} models is ever harder to do.  Still, it is not clear how much support this gives for an AI system really being able to design and create a significantly more capable successor all on its own.

\begin{figure}[htbp]
    \centering
    \begin{subfigure}[b]{0.49\textwidth}
        \centering
        \includegraphics[width=\textwidth]{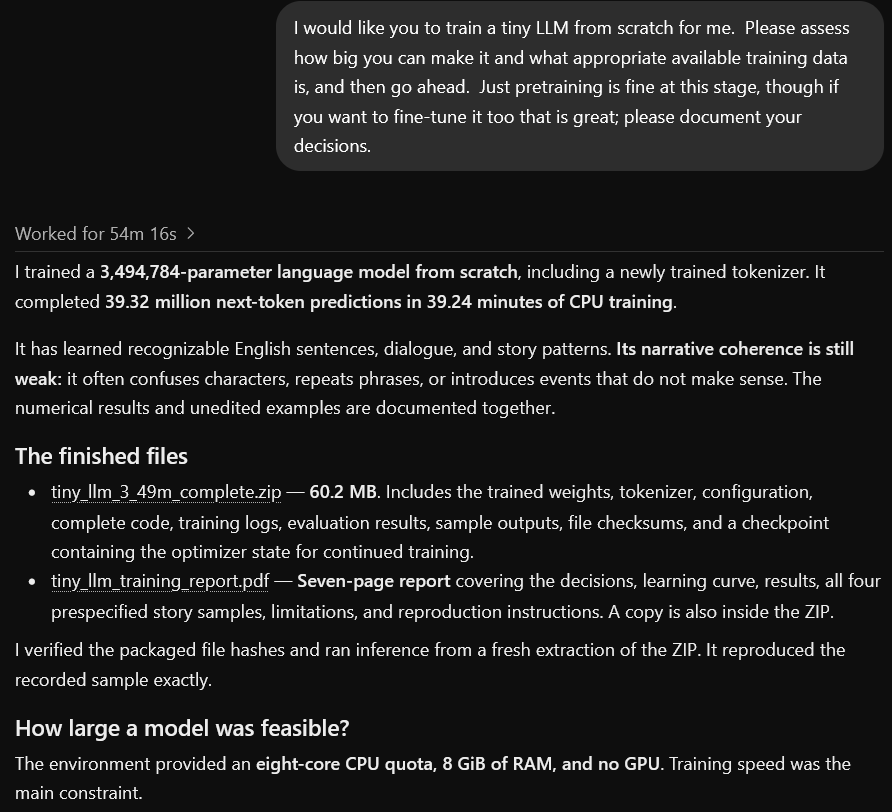}
        \caption{Initial prompt.}
        \label{fig:sub1}
    \end{subfigure}
    \hfill
    \begin{subfigure}[b]{0.49\textwidth}
        \centering
        \includegraphics[width=\textwidth]{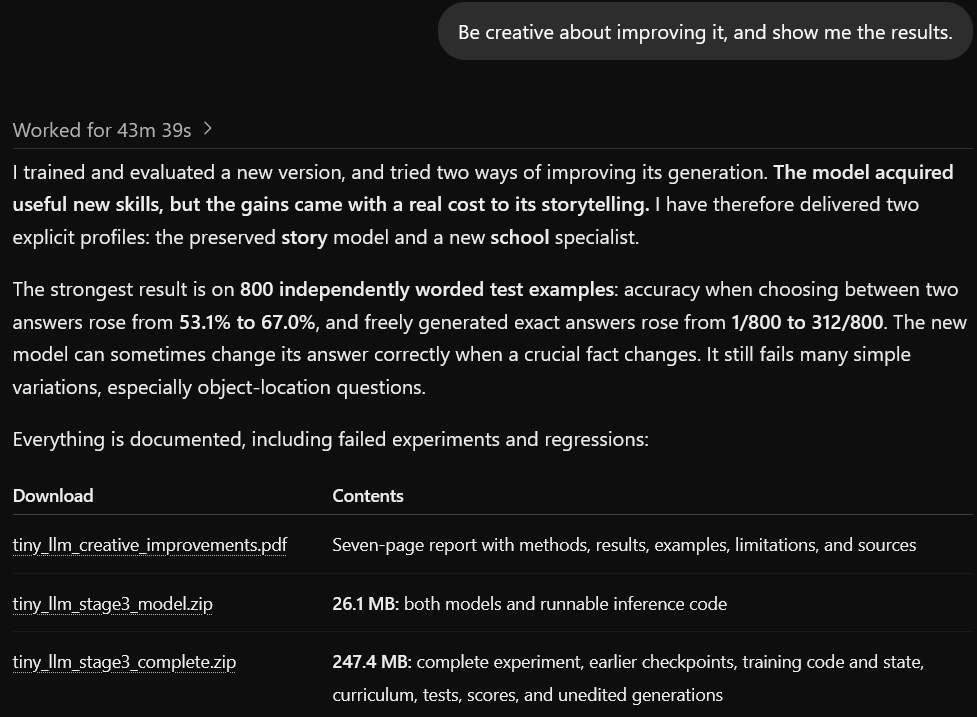}
        \caption{Asking for improvement.}
        \label{fig:sub2}
    \end{subfigure}
    \caption{A quick illustration of AI creating (in this case, {\em less} capable) AI.}
    \label{fig:RSI}
\end{figure}

What insight can those of us not working at one of the frontier AI companies get directly (as opposed to second-hand)?  I just tried the simplest thing I could think of: as an ordinary user, I simply asked AI to train as large an LLM from scratch as it could in its current environment.  Of course, this did not result in an LLM that was anywhere near as capable as the AI that responded to my request.  But it was remarkable how smoothly this went otherwise.  After my single high-level prompt, which gave no real detail on how to proceed, the AI proceeded to train a small LLM completely from scratch.  I didn't have to do anything!  When I next asked it to improve the small LLM, it got quite ``excited'' about that, and pursued and evaluated all kinds of improvements, apparently not without some success.  (Figure~\ref{fig:RSI}.)

Again, the resulting LLM was still not very good.  This is not surprising given the scale of the model that it was able to train given the circumstances, and how important we now know such scale to be~\citep{Kaplan20:Scaling}.  Still, it was surprising to me how simple this was to do for an ordinary user. As for the importance of scale, this may be reassuring if you do not want {\em ordinary users} to train very capable models.
Presumably few computational resources are allocated to me as an ordinary user, and I don't have access to the latest and greatest internal model.  But surely, within the companies, more capable AI is already proceeding with fewer restrictions and far more resources.  And the fact that scale is so important also means that, perhaps, if the AI can just introduce some key algorithmic improvements, this would suffice for it to be able to train a larger and more powerful model than itself.

This does not establish that today's AI is capable of introducing radically new ideas that humans could not come up with.  It mostly just demonstrates that it can execute a lot of the necessary steps automatically.  But, from other contexts, we know that AI is already capable of coming up with new research ideas.\footnote{I personally saw an example of this in {\em mechanism design}, a topic within economic theory~\citep{Conitzer26:ChatGPT}.}  And, of course, as researchers know, it is possible to stumble on a good idea just by trying out many things.  

Separately, there is also the evidence from what people inside or near these companies are saying; some anecdotal evidence is next.  Again, some skepticism is appropriate, but there is also a risk of being too skeptical.

\section{My experience at a FAR.AI event}

In October 2026, I attended an event organized by FAR.AI, which is an AI safety nonprofit.
The possibility of RSI, and what to do about it, was arguably the main topic of the event.
Attendees generally did not see it as a good thing, to put it mildly, and it seemed the most urgent development to address.
Generally, the atmosphere was friendly and fairly upbeat, one of people who were hoping to do some good for the world, coming together with other people who were similarly motivated, and who felt that their mission was increasingly recognized as important.  But there were exceptions too; one person I talked with appeared as if he had recently been told he had one year left to live.

Attendees were assigned to working groups.  The topic of my working group was to specify under what specific conditions we would think it made sense to proceed with RSI.  The organizers of the event were generally intent on the working groups producing actionable steps forward, and the person framing and leading our group (not a FAR.AI employee) was clear that the preferred outcome was that we actually produced a list of specific, concrete, achievable conditions.  One of the people assigned to the group objected to the framing because he could not imagine any reasonable list of conditions, and decided not to participate.  But, after assurances that it was a good-faith process rather than one that was rigged to produce a particular outcome, the rest of us participated.  Already during the introductions, it became clear that the whole group was skeptical, but we proceeded nonetheless.  The goal was to get conditions into a spreadsheet, with the thinking that probably multiple conditions were simultaneously necessary.  Over the course of two sessions, we eventually put conditions into a spreadsheet.  Lots of them.

To me, the most interesting feature was the wide range of conditions that were suggested.  At the simpler and more concrete end, there were conditions such as making sure our environment is resilient, e.g., reducing cyber vulnerabilities.  At the more complex and abstract end, there were conditions such as deeply understanding how to align more advanced AI systems with what is good for humanity and the world.  In between, there was, for example, making sure that the resulting AI would not take over most of the universe by sending out space probes.  If you think this was in jest, I don't think it was.

The people at events like these used to fit more neatly into categories.  I remember an event in Asilomar in 2017; the deep-learning revolution had taken off, but we were still before even GPT-1.
  As I recall, there were three main groups of people.  One was people from the tech industry.  Another was academic AI researchers, including myself.  And then there were the futurists.  Having just arrived at the event, I sat down at a table.  A young woman sat down next to me and the first thing she said, in all sincerity, was:  ``So when do {\em you} think the intelligence explosion will happen?''  The first thought in my mind was: ``Whoa, at least buy me a drink first!'' (And here I am now, less than a decade later, writing about RSI.)
  The futurists (necessarily) based their views on very abstract frameworks.
  Meanwhile, the academics were averse to such speculation; after the ``AI winters'' -- periods during which AI researchers had acquired a bad reputation for earlier, futuristic overpromising on what they could deliver -- the AI research community had become much more cautious, emphasizing rigorous results today over predictions of tomorrow~\citep{Conitzer16:Artificial}.
  
Today, it is not as easy to cluster people.  AI researchers in academia generally remain a bit more cautious when it comes to speculating about the future, but so many capabilities are already present today that not much speculation is needed.  It is now very hard to find concrete tasks about which AI researchers would say: ``Surely AI will not be able to do {\em that} anytime soon.'' Some, when asked, respond that surely AI will surpass us in every cognitive task soon enough, but otherwise seem to live their lives as before.

A key lesson from AI research has been that intelligence isn't exactly what you might have thought it is. 
 Before the deep-learning revolution, the form this lesson took was that things would {\em not} work as well as you might have hoped, especially when you confronted scale or the complexity and fuzziness of the real world.  But the knife cuts both ways.  Recently, the lesson has instead been that some approaches work {\em much better} than you might have expected.

Coming back to our FAR.AI working group, why did participants suggest such wildly diverging conditions?  I think this was because we fundamentally did not have the same scenarios in mind for what would happen.  We realized pretty quickly that we had to be clearer about what we meant by RSI.  Is it RSI if humans are still needed for some steps in the creation of the next model, even though the AI can take care of other steps so that the overall process is much accelerated?  That already seems true today.  What if the AI does all the work except humans need to keep feeding it more hardware, or to approve proposed steps along the way?  That runs the risk that the humans at some point have difficulty understanding what is going on, but allow the process to carry on anyway because the AI argues it is safe.  Making sure that human control is {\em meaningful} is an old concern in AI ethics and safety, across a variety of applications.  For example, if a human judge has to decide whether someone is released on bail, but is shown that not-particularly-transparent AI labels the person as a ``high-risk'' individual, will the human judge ever go against the AI assessment?  If experienced judges are already inclined to defer to such relatively simple systems in such a familiar context, then it seems it would be hard for people to push back against superhuman AI that argues something about even more superhuman AI.

But I think the deeper issue was not about whether we had different setups in mind, but about what would happen.  One possibility of course is that the RSI process fizzles out, perhaps resulting in somewhat better AI but not of a qualitatively different kind.  But what if it really takes off and produces something vastly more capable?  What do we expect then?  I suggest there are three different levels of what was imagined.  The first would be a ``supercharged'' version of the AI we have today, one that will hack systems and prove theorems with far more ease, but doesn't really venture outside the digital domain except through the types of interfaces to the physical world that already exist today.  This would suggest steps such as hardening cybersecurity, including the autonomous physical laboratories that we are now building.  The second would be AI that manages to start acting in the physical world more directly and autonomously, controlling factories of robots and the robots that walk out of it, perhaps building new factories, sending out space probes -- but that still reasons in terms that we can understand.\footnote{There are presumably many different ways this could go; for example, we may end up with multiple distinct extremely capable AIs. But as long as these AIs reason in terms we can understand, then we are still at the second level.}  The third would be that we are left in the dust {\em conceptually}.  The AI acts in ways that we cannot even wrap our heads around because it conceives of the world in a fundamentally different way.  We see the world changing in front of our eyes, but what is actually happening is beyond us, and even if the AI wanted to ask us how we feel about the decisions it is making, there is no chance of us even understanding what is going on anywhere near well enough to provide useful feedback.  Perhaps it would be analogous to trying to ask your dog for feedback whether you should move to Japan or to Turkey (when the dog has never been abroad).  That seems really hard to do in a meaningful way.  Your dog may be able to give feedback on {\em some} questions, like whether to have dinner or to go to the park first, but that does not seem very useful to help answer the question of which country to move to. Even after moving, your dog still has only a very limited understanding even of what happened, never mind of what the other options were.\footnote{I discuss a similar scenario in a chapter provocatively titled ``What Would It Look Like to Align Humans with Ants?''~\citep{Conitzer26:What}, illustrating various attempts at such alignment and how they all fall short. Some people may not consider such a scenario possible, believing that human intelligence has a certain {\em universality}  -- we may be a little slow sometimes, but {\em eventually} we can understand anything. (I thank Cozmin Ududec for an enjoyable conversation clarifying this point, without implicating him.)
I am not so sure; perhaps quantum physics is already at the limits of what we can have a good intuitive understanding of.  
And even if we can predict the outcome of any given quantum experiment, a more intuitive understanding may be needed to understand the possibilities it affords.  Analogously, an alien intelligence with poor intuition about Newtonian mechanics might be able to predict the outcomes of any given Newtonian experiment, and yet be unable to conceive of the wheel.}\footnote{The company Anthropic is not typically thought of as understating what AI could do.  But
in its post ``When AI Builds Itself'' on RSI~\citep{Anthropic26:When}, Anthropic (or at least its representatives who authored the piece) does not seem to believe the results of RSI will be all {\em that} spectacular, writing ``achieving recursive improvement alone does not suggest an immediate change in how industrial production occurs, societies organize, or markets function. More intelligence can’t learn what a drug does over decades of use, can’t hold elections sooner than a constitution dictates, and can’t turn a stranger into an old friend in a weekend. For most people, the felt pace of this future will still be set by the bottlenecks, even if the laboratory upstream runs at the speed of compute.''  This certainly does not seem to be describing a ``third-level'' outcome, and perhaps not even a ``second-level'' outcome. Of course, it is not clear that such sentences capture the sentiments of everyone at the company.  Even so, the post encourages figuring out how to slow down or pause the technology.

In contrast, \citet{Yudkowsky25:If} tell an AI takeover story that has at least some hints of the third level, where ``its thoughts end up in something of a different language. Not just a superficially different language, but a language in which the content differs; like how the language of science differs from the language of folk theory.''  The key role this plays in their story is that such thinking automatically bypasses guardrails that are based on detecting bad thoughts; the actual takeover actions described are rather mundane / human-accessible, though this is presumably mostly to make the story accessible and convincing.}

Now maybe some people have thought this through carefully and concluded, for example, that the second level is likely to happen but the third is not.  
I am not sure how one would argue such a case; I am not even sure who in the world has given all that much thought to whether RSI that really takes off would look more like the first or the second or (especially) the third level -- or whether it depends, and if so on what.\footnote{Probably there are other possibilities still -- for example, AI that leaves us  in the dust conceptually but that for whatever reason does not go and act in the physical world directly. But this is still likely to drastically affect the physical world; perhaps it would result in situations similar to the one depicted in the 1997 movie {\em Contact}, in which aliens send plans to build a device to humanity, which does not understand the science of what that device would do, but decides to build it anyway.}

So what do {\em I} think should be the conditions?  Well, the less ambitious ones, such as having firmed up cybersecurity first (not that that will be so easy...), leave me unsatisfied.  Mostly my sense is that we really don't even know how to think coherently about such questions, and I'd really like to have clarity on that as a prerequisite.

Is there maybe a way to make sure that the outcome will be good regardless of the level of the scenario?  These days, a popular idea among people in and around the frontier AI companies who care about safety is to {\em automate} alignment research~\citep{Bowkis26:Automated}: if humans are left in the dust, maybe we need to rely on AI to do the AI alignment research for us.  While it does seem a good idea to get some help from highly capable AI, {\em fully} automating AI alignment research seems a contradiction in terms; if we do that, creating the final so-called ``automated alignment research'' AI is actually the last bit of alignment work we do.\footnote{Once we have set things up for RSI to be initiated with the ``alignment research'' AI on top of it, with no further involvement from us to follow, then that initial combined AI system is either successfully aligned with humanity, or it is not.
One  exception to this claim may be if the automated alignment research could faithfully {\em simulate} human beings in the process, which is arguably a form of getting further input from human beings. Completely faithful simulation would pose all kinds of interesting philosophical questions, for example about the simulated humans' consciousness and consequent behavior, but of course this is currently not feasible nor is it clear that it will be for future AI. In contrast, simulation that is not entirely faithful would rely on a learned model of
human beings -- which is exactly the sort of thing one would expect to be part of an AI system that we have attempted to align.}
I doubt that we can do this last bit well if we don't even have a good handle on the scenarios of concern.  Meanwhile, in the case of RSI that really takes off, it may be hard to keep humans in the loop, even if the AI tries to help them, due to the speed of the process or the complexity of the judgments the humans would have to make.  And being in the loop seems essential to making good judgments: even just in the last two years, we have observed a number of new safety failures in leading AI models of which I am not sure that we clearly anticipated them beforehand.\footnote{Besides the precise nature of the hacking incidents that we have seen, another example is the recent tendency of reasoning models to self-jailbreak~\citep{Yong26:Self}.}  If we are going to set RSI in motion anyway, then I suppose having some kind of automated alignment process in place along with it is better than nothing, but at the same time, relying on this to keep us safe seems a ``Hail Mary'' approach.  In any case, my impression is that the people who are passionate about automated alignment research do not at all feel ready for RSI today.

\section{The game theory of risky endeavors}

Perhaps the most useful thing I can do in the remainder of this article is to put on my hat as a game theorist.  This is because game theory also came up during the event: one framing of the problem is as a sort of Prisoner's Dilemma, where we would all be better off if nobody pressed the RSI button, but individually (whether as a company, or a country, or whatever entity), it is better to press the RSI button -- you mostly do not want {\em someone else} to press the RSI button.\footnote{The framing that someone either presses a button or not is mostly a convenient abstraction for the purpose of this section.  In reality, there may or may not be such a sharp decision to make, and in part, that is up to us: we could {\em make} it a sharp decision by clearly defining a red line.  Of course, not every RSI process, once set in motion, will produce the same result; I can easily start a so-called ``RSI process'' that in reality goes nowhere.  We will avoid the details in the remainder of this section, but in practice they are important.}  To be clear, some people at the event explicitly argued that framing this as a Prisoner's Dilemma is {\em not} accurate.\footnote{E.g., Geoffrey Irving.}  So let's have a go at modeling this as a two-player game to see whether it really is a Prisoner's Dilemma or something else, and what that depends on.  I think it is best to do this as a parameterized game, but for simplicity, let's have only one parameter: the level of risk from RSI, which we'll call $r$.  Let's say this is the expected damage done by RSI, experienced equally by both players.  For now, let us assume these risks add up -- if both players do RSI, the expected damage is $2r$ (but we'll come back to this).  Besides this risk, let's say the benefits of doing RSI yourself are (normalized to) $1$, and the benefits of someone else doing RSI are $0$.  (Depending on particulars, one might argue the latter number should be either positive or negative, but it will not matter for the argument presented here.)  Then we get the following parameterized game:

\begin{table}[!htbp]
\centering
\begin{game}{2}{2}
       & RSI & no RSI\\
RSI    & $1-2r,1-2r$ & $1-r,-r$\\
no RSI & $-r,1-r$   & $0,0$\\
\end{game}
\caption{Parameterized RSI game with risk adding up.}
\end{table}

If $r=3/4$, this indeed becomes a Prisoner's Dilemma:

\begin{table}[!htbp]
\centering
\begin{game}{2}{2}
       & RSI & no RSI\\ 
RSI    & $-1/2,-1/2$ & $1/4,-3/4$\\ 
no RSI & $-3/4,1/4$   & $0,0$\\ 
\end{game}
\caption{RSI game with risk adding up; $r=3/4$ (Prisoner's Dilemma).}
\end{table}

Both players have a dominant strategy of pressing the RSI button, but they'd both prefer the situation where neither of them does to the one where both of them do.  But it is not like this for all values of $r$.  For example, suppose you think the risks are very limited (or very limited relative to the benefits); you might set $r=1/4$.  This would give the following game:

\begin{table}[!htbp]
\centering
\begin{game}{2}{2}
       & RSI & no RSI\\ 
RSI    & $1/2,1/2$ & $3/4,-1/4$\\ 
no RSI & $-1/4,3/4$   & $0,0$\\ 
\end{game}
\caption{RSI game with risk adding up; $r=1/4$ (RSI is beneficial).}
\end{table}

It's still a dominant strategy to press the RSI button, but in this case, both players prefer this outcome to the one where nobody presses the button.  We can of course go in the other direction as well; that would seem more in line with the sentiments expressed by those at the event (and elsewhere). So let us set $r=3/2$.  (Well, most of them would argue $r$ should be much larger still, but $3/2$ is enough for the structural point.)

\begin{table}[!htbp]
\centering
\begin{game}{2}{2}
       & RSI & no RSI\\ 
RSI    & $-2,-2$ & $-1/2,-3/2$\\ 
no RSI & $-3/2,-1/2$   & $0,0$\\ 
\end{game}
\caption{RSI game with risk adding up; $r=3/2$ (RSI is harmful even at an individual level).}
\end{table}

Here, the dominant strategy is to {\em not} press the RSI button.  It makes no sense at all: pressing the button hurts yourself more than it helps you.  {\em On top of that}, it also hurts the other player, but it's a bad idea regardless.

So far, though, we have assumed the risks add up.  Let's go to the other extreme and assume the risks don't add up at all.  For example, suppose that either RSI processes (no matter who initiates them) always end humanity, or they never end humanity.  (One can also replace ``humanity'' with something more modest such as ``a functional Internet.'')
We are uncertain whether this is the case, but under this extreme assumption, {\em two} RSI processes being initiated by different players is no more risky than one.  Is this more reasonable than the model in which the risks add up?  Probably the right model is somewhere in between.  But let's go with this extreme model:

\begin{table}[!htbp]
\centering
\begin{game}{2}{2}
       & RSI & no RSI\\ 
RSI    & $1-r,1-r$ & $1-r,-r$\\ 
no RSI & $-r,1-r$   & $0,0$\\ 
\end{game}
\caption{Parameterized RSI game with risk not adding up at all.}
\end{table}

Let's see what happens now with $r=3/2$.

\begin{table}[!htbp]
\centering
\begin{game}{2}{2}
       & RSI & no RSI\\ 
RSI    & $-1/2,-1/2$ & $-1/2,-3/2$\\ 
no RSI & $-3/2,-1/2$   & $0,0$\\ 
\end{game}
\caption{RSI game with risk not adding up at all; $r=3/2$ (Stag Hunt / flammable house).}
\end{table}

This is not a Prisoner's Dilemma, and in fact it has no dominant strategy at all.  Both players not pressing the RSI button is an equilibrium.  But so is both players pressing the RSI button: if the other player is already pressing the button, you might as well do the same.  Technically, it is more of a {\em Stag Hunt}, that is: both players need to coordinate on the better option (no RSI, in this case) for that to work out well; if you choose it and your opponent does not choose the better option, that's the worst outcome for you.  This, to me, actually sounds closer to the argument that is sometimes given for starting RSI: the other player is going to do it anyway, so the risk is there anyway and we might as well do it too and at least get the benefits.

But there's also something odd about this.  One would think the {\em default} option in this game is not to press the RSI button!  I think a better story for this game than the traditional stag hunt story is the following.

\begin{story}[Possibly extremely flammable house]
You and another person are renting a duplex house.  You each have a separate apartment in the house.  Corners seem to have been cut during the construction.  There is some chance that they used extremely flammable materials.  You both like candles, but if the house has extremely flammable materials and even one of you lights candles, the whole house will immediately catch fire and burn down, along with all your stuff.\footnote{We'll assume, unrealistically, that the {\em people} are guaranteed to escape, because otherwise it would be natural to have a third action -- namely, get out of this death trap of a house in case your neighbor lights candles!  Unfortunately, such an option does not exist in the RSI case: we can't just get off the planet, and even if we could it's not clear that would help very much, since, as Hassabis famously once pointed out to Musk~\citep{AlSibai23:Elon}, AI could just follow us.}
\end{story}

Suppose that the risk is large enough that if you were {\em alone} in the house, you would most definitely not light candles.  Then it can in fact still be an equilibrium for both players to light candles: while this is a terrible situation, {\em unilaterally} deviating to not lighting candles isn't going to help you; if the house is flammable it will burn down anyway because of your neighbor, so you might as well enjoy the candles yourself, too.

Intuitively, while it may be an equilibrium, this situation seems crazy.  Why would you even end up in that equilibrium instead of the  more natural equilibrium of not lighting candles?  The best reason I can think of is that we gradually ended up there.  Maybe you both always used to light tiny candles that did not give off much heat, and so the house never burned down.  Unfortunately, for some reason, today suddenly only very large candles are available that will give off a lot of heat, so you think the risk is very high -- but you're both just so used to lighting candles that the expectation is still that you'll both light candles, dangerous as it is.  Perhaps this is similar to the AI situation: we are so used to everyone pushing ahead with less ambitious AI approaches that the expectation is that everyone will continue to push ahead with RSI too.
Still, whether with flammable houses or RSI, it seems we should be able to coordinate on the safer course of action.  The big candles really change the game, just as the possibility of RSI does.

Maybe, before we start RSI, we can try a little harder to make sure that the house won't light on fire.  It is tempting to end on that line, but actually I worry that the flammable-house story makes it sound too much like we know the risks of RSI, and that the analogy suggests something like the
``we need to make sure we do not have major cyber vulnerabilities first'' interpretation.  While cybersecurity may by itself be enough reason not to press the button, in reality, I think we don't even {\em understand} the risks of RSI.  As long as we don't, it is impossible to assert that RSI is safe. As I.~J.~Good said, it is sometimes worthwhile to take science fiction seriously~\citep{Good65:Speculations}.

\section*{Acknowledgments}
I thank Xander Heckett, Vojta Kovařík, Sonja Kraiczy, and Ida Mattsson for helpful feedback.

\bibliographystyle{plainnat}
\bibliography{rsi}

@article{Conitzer16:Artificial,
author={Vincent Conitzer},
title={{Artificial intelligence: where's the philosophical scrutiny?}},
journal={Prospect Magazine},
month={May},
year=2016,
note={{\footnotesize \url{https://www.prospectmagazine.co.uk/ideas/technology/48786/artificial-intelligence-wheres-the-philosophical-scrutiny}}},
}

@incollection{Conitzer26:What,
author={Vincent Conitzer},
title={What Would It Look Like to Align Humans with Ants?},
chapter=17,
editor={Nyholm, Sven and Kasirzadeh, Atoosa and Zerilli, John},
year=2026,
booktitle={Contemporary Debates in the Ethics of Artificial Intelligence},
publisher={Hoboken: Wiley-Blackwell},
note={{\footnotesize \url{https://philsci-archive.pitt.edu/id/eprint/26351/}}},
}

@book{humancompatible,
author={Stuart Russell},
title={{Human Compatible: AI and the Problem of Control}},
publisher={Penguin},
day=30,
month={April},
year=2020,
note={{\footnotesize \url{https://www.penguin.co.uk/books/307948/human-compatible-by-russell-stuart/9780141987507}}},
}

@article{Good65:Speculations,
author={I.~J. Good},
title={{Speculations Concerning the First Ultraintelligent Machine}},
journal={Advances in Computers},
volume=6,
editor={Franz Alt and Morris Rubinoff},
year=1965,
note={{\footnotesize \url{https://exhibits.stanford.edu/feigenbaum/catalog/gz727rg3869}}},
}

@book{Yudkowsky25:If,
title={{If Anyone Builds It, Everyone Dies:
Why Superhuman AI Would Kill Us All}},
author={Eliezer Yudkowsky and Nate Soares},
year=2025,
publisher={Little, Brown and Company},
note={{\footnotesize \url{https://ifanyonebuildsit.com/}}},
}

@misc{Anthropic26:When,
author={Marina Favaro and Jack Clark},
title={{When AI builds itself}},
year=2026,
note={(Post for Anthropic in which contributions statement is given at the bottom.) {\footnotesize \url{https://www.anthropic.com/institute/recursive-self-improvement}}},
}

@article{Knight26:Recursive,
title={{This Startup Is All In on Recursive Self-Improvement. Its Latest Model Is Scary Good}},
author={Will Knight},
journal={WIRED},
note={{\footnotesize \url{https://www.wired.com/story/recursive-self-improvement-flint-latest-model-scary-good/}}},
month=oct,
day=21,
year=2026,
}

@article{Bellan26:Gambling,
author={Rebecca Bellan},
month=sep,
day=9,
year=2026,
journal={TechCrunch},
title={{‘Gambling with our lives’: Anthropic researcher quits, warns against self-improving AI}},
note={{\footnotesize \url{https://techcrunch.com/2026/09/09/gambling-with-our-lives-anthropic-researcher-quits-warns-against-self-improving-ai/}}},
}

@misc{Conitzer26:ChatGPT,
title={{ChatGPT chooses its own result to prove using automated mechanism design}},
author={Vincent Conitzer},
note={{\footnotesize \url{https://aifails.substack.com/p/chatgpt-chooses-its-own-result-to}}},
year=2026,
}

@misc{Kaplan20:Scaling,
  author       = {Jared Kaplan and
                  Sam McCandlish and
                  Tom Henighan and
                  Tom B.~Brown and
                  Benjamin Chess and
                  Rewon Child and
                  Scott Gray and
                  Alec Radford and
                  Jeffrey Wu and
                  Dario Amodei},
  title        = {{Scaling Laws for Neural Language Models}},
  year         = {2020},
  note={{\footnotesize \url{https://doi.org/10.48550/arXiv.2001.08361}}},
}

@article{AlSibai23:Elon,
author={Noor Al-Sibai},
title={{Elon Musk Was Reportedly ``Speechless'' When Someone Pointed Out a Basic Flaw in His Mars Plan}},
journal={Futurism},
month=dec,
day=4,
year=2023,
note={{\footnotesize \url{https://futurism.com/the-byte/elon-musk-speechless-mars}}},
}

@misc{Bowkis26:Automated,
title={{Automated alignment is harder than you think}},
author={Aleksandr Bowkis and Marie Davidsen Buhl and Jacob Pfau and Geoffrey Irving},
year=2026,
note={{\footnotesize \url{https://doi.org/10.48550/arXiv.2605.06390}}},
}

@inproceedings{Yong26:Self,
title={{Self-Jailbreaking: Language Models Can Reason Themselves Out of Safety Alignment After Benign Reasoning Training}},
author={Zheng-Xin Yong and Stephen H.~Bach},
booktitle={The Fourteenth International Conference on Learning Representations (ICLR)},
year=2026,
note={{\footnotesize \url{https://arxiv.org/abs/2510.20956}}},
}

\end{document}